\documentclass[conference]{IEEEtran}
\IEEEoverridecommandlockouts
\usepackage{cite}
\usepackage{amsmath,amssymb,amsfonts}
\usepackage{graphicx}
\usepackage{textcomp}
\usepackage{xcolor}

\def\BibTeX{{\rm B\kern-.05em{\sc i\kern-.025em b}\kern-.08em
    T\kern-.1667em\lower.7ex\hbox{E}\kern-.125emX}}
\begin{document}

\title{A Novel Framework for Neural Architecture Search in the Hill Climbing Domain
\\
}


\author{\IEEEauthorblockN{Mudit Verma, Pradyumna Sinha, Karan Goyal, Apoorva Verma and Seba Susan}
\IEEEauthorblockA{Department of Information Technology \\ Delhi Technological University\\
Bawana Road, Shahbad Daulatpur, Rohini, Delhi, 110042\\
Email: \{mudit.verma2014, pradyumna014, karangoyal2k96, 19apoorva.v\}@gmail.com, sebasusan@dce.ac.in
}
}

\maketitle

\begin{abstract}
Neural networks have now long been used for solving complex problems of image domain, yet designing the same needs manual expertise. Furthermore, techniques for automatically generating a suitable deep learning architecture for a given dataset have frequently made use of reinforcement learning and evolutionary methods which take extensive computational resources and time. We propose a new framework for neural architecture search based on a hill-climbing procedure using morphism operators that makes use of a novel gradient update scheme. The update is based on the aging of neural network layers and results in the reduction in the overall training time. This technique can search in a broader search space which subsequently yields competitive results. We achieve a 4.96\% error rate on the CIFAR-10 dataset in 19.4 hours of a single GPU training.
\end{abstract}


\section{Introduction}
Deep Neural Networks have demonstrated impressive performance on
many machine-learning tasks such as computer vision, speech recognition and natural language processing \cite{b1}\cite{b2}\cite{b3}\cite{b4}. The high performance of deep neural networks is strengthened by the cost of large-scale engineering and checks to find the best architecture for a given problem. High-level design decisions such as depth, units per layer, and layer connectivity are not always obvious and use hit and trial approach which is exhausting and time-consuming. This led to a growing interest in the automatic designing of neural networks. Since the architecture search space is discrete, traditional optimization algorithms such as gradient descent\cite{b20} are not applicable, hence approaches based on reinforcement learning algorithms\cite{b13} and evolutionary algorithms\cite{b21} are used for automated designing of neural networks. But these approaches are either very costly (requires hundreds and thousands of GPU days) or yield non-competitive results.
In this work, we aim to reduce the computational cost by using a simple hill-climbing approach for the optimization of search space and our variant of network morphism\cite{b18} for making the network deeper and complex. We have also added gradient stopping, linear morphism and several implementational improvements for reducing the training time.
Our work uses a single GPU, and we validate our method on the CIFAR-10 dataset. It takes our framework 19.4 hours to reach 4.96\% error. When we use this same architecture to train and validate on MNIST dataset, we obtain an error percentage of 0.28. We have compared two ways of weight initialization, with and without Gradient Stopping.

In the following sections, we will discuss the previous attempts at Neural Architecture Search (NAS). Further, we present our NASGraph, explain its various nodes and morphism operations, and propose our methodology. We describe our implementation of morphism operations, gradient stopping technique, linear morphism and finally give a flow chart for our algorithm. In the end, we present our experiments and showcase our results in comparison with handcrafted architectures on CIFAR-10 as well as other Neural Architecture Search schemes, and also gain insights about using our generated deep neural network on different datasets. We then conclude our work and present a few pointers for possible future research opportunities based on our work.

\section{Related Work}

Research on neural architecture search started in the 1980s. Initially, evolutionary algorithm based approaches were proposed which promises to find out both the architecture and weights of the neural network in \cite{b5}. However, they have been unable to match the performance of the hand-crafted network. More recent work \cite{b6}\cite{b7}\cite{b8}\cite{b9}, use an evolutionary algorithm only to find out the structure and stochastic gradient descent to find out the parameters of the network. \cite{b9} and \cite{b10} use evolutionary algorithms to generate a complex architecture from a smaller neural-network iteratively, but \cite{b9} used a large number of resources (250 GPUs and ten days) whereas \cite{b10} is limited to relatively smaller networks due to handling a population of networks.

\textbf{RL-based approach.} \cite{b12} proposed Neural Architecture Search (NAS) with the help of reinforce algorithms\cite{b13} to learn a network architecture. It generates a sequence of actions representing the architecture of a CNN. NASNet\cite{b14} uses proximal policy-optimization (PPO) in-place of reinforce algorithms\cite{b13} which further improves neural architecture search. It searches the architecture of a small unit which they referred to as “block.” It repeatedly concatenated itself to form a complete model. Other works in the field use approach like policy gradient in\cite{b14}, Q-learning in\cite{b15} for finding the model architecture. Unfortunately, training a reinforcement learning agent for searching architecture is extremely expensive: both \cite{b16} and \cite{b17} required over 10.000 fully trained networks, requiring hundreds to thousands of GPU days.

\textbf{Network morphism/ transformation :} Using the concept of transfer learning, network morphism was introduced and first published in 2015 by Chen \cite{b18}. To make the network more complex, the author described function preserving operation such as 'deeper' or 'wider' with the goal of speeding up training and exploring network architectures. \cite{b18} further proposed additional operations, e.g., for handling non-idempotent activation functions or altering the kernel size and introduced the term network morphism. \cite{b19} presents a neural architecture search by hill climbing (NASH) and used morphism operations for making the network more complex.
Out of all the methods, \cite{b19} resembles the most to our techniques of neural architecture search by hill climbing (NASH) with network morphism. We extend their work by adding gradient stopping, linear morphism, several implementational improvements like the selection of nodes upon which morphism is to be applied, the addition of more number of operators for growth of performance using less resources, time (1 GPU and less than one training day) and at the same time yielding higher accuracy.

\section{NAS Graph}
\label{NASGraph}
NASGraph represents an instance of neural-architecture in our state space of neural network architectures for the dataset that is given to the hill climbing algorithm. A specific set of attributes define a NASGraph. It is implemented as a Graph data structure where each node can either be a convolutional block (ConvLayer-BatchNorm-ReLU), a max pool node, a Merge node or a linear node. It has an adjacency list, the matrix that defines the connections between the nodes. It also stores a list of node identifiers and a mapping between the identifiers and the actual instances. There is another list that maintains the node identifiers of its parents and children. Morphism operations have been defined on the NASGraph, and during the hill climbing process, they are called to "morph" the parent network into a child network. At an abstract level, the graph is initialized with image (from the input dataset) dimensions. A Convolution node is added with random parameters initially, and linear layers are stacked at the end for classification. Finally, the NASGraph is transformed into a network, and its parameters are provided to the SGDR\cite{b23} optimizer.

\textbf{Topological Ordering Condition:} According to this condition, it is a must for the NASGraph to remain acyclic at all times. Therefore any connections made from node $x$ to $y$ in the NASGraph must be such that $x$ lies before $y$ in the topologically sorted order of the NASGraph.

\subsection{NASGraph Nodes}

NASGraph Nodes are the fundamental building blocks of all possible neural architectures through our framework. These nodes have dual behavior; they are the usual graph nodes and therefore participate in lists maintained for graph representation. However, they are also network layers; this limits the position of edges in the graph. Also, to implement the framework, algorithms such as topological sort, cycle detection, etc. have been performed on these which are used later to maintain the topological ordering condition.

There are a total of 4 NASGraph Nodes as follows.
\begin{enumerate}
    \item \textbf{Convolution :} This node is composed of a 2-D convolution layer followed by a Batch Normalization\cite{b24} layer and ReLU activation.
    
    \item \textbf{Maxpool :} This node is composed of a max pool layer.
    
    \item \textbf{Merge :} This node may be an Add layer, which takes up multiple node outputs as it's input, where all inputs must be of the same dimension and output their matrix addition. Additionally, this node may be a Merge layer, where different inputs are stacked one below another along channel dimension and fed as one input to the next layer.
    
    \item \textbf{Linear :} This is a fully connected neural network layer.

\end{enumerate}

The first node of the NASGraph is supposed to be randomly created, and after this, the iterative hill climbing begins. The first node is a convolution node with kernel sampled uniformly from set $\{3,5\}$, padding, uniformly from set $\{True, False\}$, $stride = 1$, and channels from set $\{8,16,32\}$. Beyond this point, the only way to modify the properties of an existing NASGraph node is via morphism operations.

\subsection{NASGraph Morph Operations} \label{NASGraph Morph Operations}

\begin{figure}[htbp]
\centering
\includegraphics[width=0.45\textwidth]{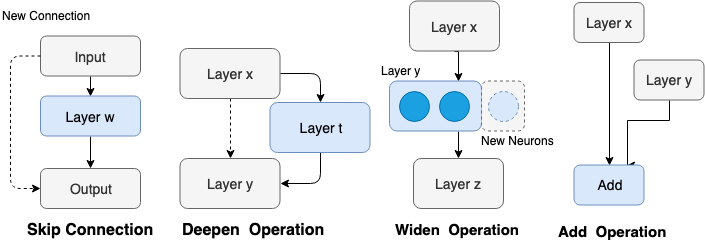}
\caption{Some of the NASGraph Operations} \label{operations}
\end{figure}

Our morphism technique differs from that of \cite{b19} in that we do consider layers without padding. Convolution layers without padding have proven to be useful as seen in popular architectures such as VGG16\cite{b25}. Therefore, this factor must be included in the search space. 

We allow our NASGraph to perform one of the following six operations for the generation of the child node in the hill climbing algorithm :

\subsubsection{Skip Operation}

Skip Operation allows the NASGraph to have skip connections in the neural network. Skip connections help traverse information in neural networks. As the NASGraph is found to generate very deep neural architectures, it is possible for the gradient information to be lost since it passes through so many layers, thus skip connections enable us to pass feature information to lower layers to help classify directly. As we also provide the provision for max-pooling layers, max pooling operation itself causes some amount of spatial information to be lost, skip connections help in this as well.

The idea for performing Skip is first to find all the node pairs in which the first node’s output is the same as the second node’s input. Such a situation may cause the second node to have multiple parents each, providing this node with same dimension inputs. Therefore, we use an Add node, described below, to add all such parents and then supply the combined result to this node in Fig. \ref{operations}. We can see an example of a skip connection being applied between the input and output layer shown by the dashed line connection.

\subsubsection{Deepen}

Deepen, as the name suggests, is a layer to deepen the existing network. This layer, in our case, is a Convolutional Block composed of a Convolutional Neural Layer, a Batch Normalization Layer, followed by a ReLU activation function. The layer may also be a Linear Layer, that is a fully connected layer. 

To perform the deepen operation between nodes $x$ and $y$, let us assume, layer $x$ has $a$ neurons, and layer $y$ has $b$ neurons. Then, the weight matrix is of shape (a,b). The new layer $t$ must have input dimension of (b,b) so that the composition of layers $x$ and $t$ gives the final output (a,b), same as before.

Selection of Nodes for Deepen is given in section \ref{Selection of Nodes For Morphism}. Figure \ref{operations} shows a new node $t$ being added in between nodes $x$ and $y$. The addition of a layer adds another level of abstraction that cannot be as simply contained within shallow layer architecture of the same number of parameters if they can be contained at all.

\subsubsection{Widen}

Widen operation increases the channels of a convolution node by a factor of 2 or 4. Widening requires changing the parent node weight matrix as well as the subsequent child node weight matrix to adjust for the widening changes. To simplify the process, we, therefore, create a set of those node pairs, in their topological ordering, $(x,y)$ where has only one child $y$, and $y$ has only one parent $x$. Not using this condition would require us to perform changes in the architecture recursively, i.e., if the parent weight changes, then all its children weight matrices have to change recursively.

Widen Operation has also been used for fully connected layer widen morphism which is described in detail in section \ref{Linear Morphisms}.

\subsubsection{Add}
When a node in the graph has more than one parent, one of the options to combine the inputs is to do an element-wise sum of the incoming tensor. However, this is only possible if the incoming tensors are of the same shape. 

\subsubsection{Merge}
The other option to combine inputs from more than one parent is to concatenate the channels of the image tensors. For this method to be used, it is required that the height and width of the incoming image tensors be the same.

\subsubsection{Maxpool}
The max pool property is essential to building CNN and is used extensively in object detection tasks like in \cite{b25}. \cite{b19} does not allow for dynamic addition of new max-pool nodes in the architecture, i.e., max pool layer must be originally present in the seed architecture for it to exist in the final architecture. We ease this restriction and allow for dynamic addition of max pool nodes in our NASGraph. Similar to the addition of a Convolution Node, we first find all node pairs, in topological order, which can accommodate a max pool layer in between them and maintain all previous connections. Then, we uniformly sample one such node pair and add a max pool.

\section{Proposed Work}

\subsection{Morphism}
\label{Morphism}
Morphism has been used in\cite{b19} before for Neural Architecture Search; however, in approaches like NASH\cite{b19}, the use of Morphism restricted the network state space, i.e., all Convolution Layers are required to be padded. They treat each convolution layer as a function, which they split into two functions using the morphism operation, thereby the final input and output dimensions remain the same as before the morphism was applied. We, however, allow layers to be added without padding, which dynamically changes the output dimension of the layer. 

Morphism functioning for different operations has been explained in Section \ref{NASGraph Morph Operations}. The morphism has been applied in two ways, first with default Pytorch version 0.4.1 \cite{b28} initialization of layer weights (Xavier Initialization for Convolutional Layers), and the second, 0/1 initialization as in NASH. In 0/1 initialization when we use widen operation, the added neurons are set to value 0, and for deepening process, the added layer has all values initialized to 1. We show our results with both types of initialization methods and find that they do not affect the total time to get a trained network, and default initialization performs better than 0/1 initialization only marginally.

\subsection{Selection of Nodes For Morphism}
\label{Selection of Nodes For Morphism}

NASH\cite{b19} states the morphism technique and the operations; however, the selection of the nodes upon which the morphism operation would take place is described as "random." This random selection works for the NASH procedure because they begin with a seed architecture and each added layer has padding enabled, thereby keeping the output always the same. In our case, however, compatibility of nodes between which an operation is to be performed is an issue. For example, in the Skip operation, we need to pick two nodes A and B such that, $output(A) = input(B)$. Moreover, any such additional connection must meet the topological order condition of NASGraph as in section \ref{NASGraph}. 

We always keep a topological ordering of the NASGraph nodes, and for any request of connecting two nodes via Skip, compatible node pairs ordered topologically are found, and one such pair is selected uniformly randomly. This method of 'randomly' picking nodes for the morphism operation can be termed as fairly random for all practical purposes because it is essentially selecting nodes for the operation randomly from the set of all compatible nodes. 

For adding a Convolution Layer with a given kernel shape and other parameters, say c, we find a set of all topologically ordered node pairs (x,y) such that $c(x.output).output = y.input$. Now we can randomly pick any element of this set for our morphism operation.

\subsection{Random Selections}
\label{Random Selections}

Many parameters of the NASGraph nodes are sampled randomly and we describe all such parameters below.

We need to select various properties of the NASGraph nodes, for example, the kernel, stride, etc. of the Convolution Node. As in \cite{b19}, we choose these properties randomly from a predefined set of values. And finally, since we allow some of the layers to exist without padding, we mark padding True or False, with a probability of $0.5$ each.

For deepen operation which is the addition of a Convolution Node, we select the kernel from the set $\{3,5\}$. The number of channels is chosen to be equal to the number of channels of the closest preceding convolution.

For widen operation which is the expansion of a Convolution Node, the number of channels increases by a widening factor, which is sampled randomly from the set $\{2,4\}$.

The parameters for Maxpool operation are also required to be chosen uniformly. Kernel for the same is selected uniformly from a set $\{2,3\}$. 

The morphism operation applied is also chosen randomly from the set of all possible morphisms. We describe morphism operations in the section \ref{NASGraph Morph Operations}. Finally, if a certain operation cannot be performed at some point of time, then we temporarily remove that operation from the set and again sample uniformly from this new operations set.

\subsection{Gradient Stopping}

\begin{figure}[htbp]
\centering
\includegraphics[width=0.45\textwidth]{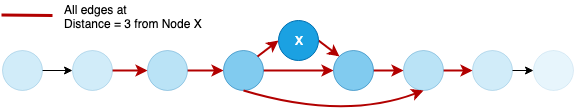}
\caption{Illustrating the nodes involved in gradient stopping for distance = 3} \label{edge}
\end{figure}

Repeated application of NASGraph morphism operations for a large number of times can result in very deep neural networks. The problem with the conventional approach to handling these dynamically growing neural architectures is overtraining of specific 'old' layers. Additionally, we try to analyze the addition of a layer via a water drop ripple analogy. A drop of water falling in a pool of water creates water ripples around the drop location, thus leaving far away locations unaffected. Hence, the addition of a convolution layer should effectively cause major changes in the weights of near-by layers only, and training of the far-away layers may be stopped. However, defining a set of nodes as 'far away' is difficult, so we use a Normal Distribution Curve to solve this.  

By iterative generation of complete Neural Networks, we found the average depth of the graph is around 36. We arbitrarily choose to let the training of nodes continue which lie within $1/3$ of the average length of nodes which comes to be approximately 12. Thus after adding a convolution node, we find all the nodes connected up till distance = 6 (which would give at least 12 nodes as the total from either side). An example of distance = 3 is presented in Fig. \ref{edge} where all the edges at a distance less than or equal to three have been marked red. Now, all the nodes connected via these marked red edges can be denoted via their distances.

To determine the learning rate of a node, a value from the Normal distribution equation 
$ f(distance) =  \frac{6}{\sqrt{2\pi}\cdot 2.4}e^{\frac{-\left(x\right)^2}{2s^2}}$ corresponding to the distance assigned to the node is obtained, say $alpha_n$. The distribution is set so that this $alpha_n$ lies between the range $(0,1]$ and tends to zero as the distance value reaches $6$. Beyond $distance = 6$ we set the value of $alpha_n = 0$. Now, we obtain the learning rate value $alpha$ according to the SGDR\cite{b23} update rule and get the new $alpha = alpha*alpha_n$. Note that in the next iteration, $alpha$ is again obtained from SGDR rule, as it would be without Gradient Stopping, and the changes are performed on that value. All the nodes beyond distance 6 do not update their gradient until the next non-layer-addition morphism operation.

\begin{figure}[t]
\centering
\includegraphics[width=0.45\textwidth]{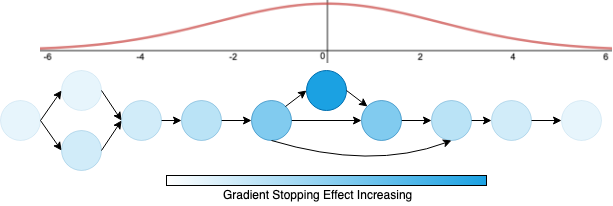}
\caption{Gradient Stopping, using the normal distribution to update new learning rates for neural network nodes.} \label{gradstopping}
\end{figure}

Fig. \ref{gradstopping} gives a graphical representation of gradient stopping, where the darkest circle represents the newly added node, for which the learning rate is unchanged, as $alpha_n = 1$. As we move away from this node, the circle shade fades off representing reduction in their learning rates, and beyond a specific number of nodes (6 in this case), the training completely stops. Gradient Stopping is a considerably good example of layer-wise training of a neural network with dynamic changes to the architecture at the run time.

\subsection{Linear Morphisms}
\label{Linear Morphisms}
\begin{figure}[h]
\centering
\includegraphics[width=0.45\textwidth]{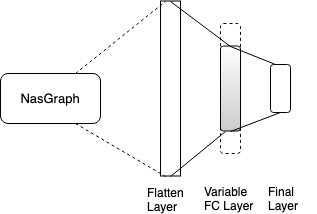}
\caption{Linear Widening Morphism} \label{linearMorphism}
\end{figure}

As mentioned previously, our framework supports convolution nodes to be added without padding, as well as the addition of max pool layers. The output dimensions of these layers are strictly less than the input dimensions; therefore, the final output shape of tensor exiting the 'NasGraph' (graph of Convolution, MaxPool, Add, Merge Nodes) bubble reduces. For example, if the previous output dimension is (28,28), now with the addition of Convolution Node at the very end with kernel size (5,5) without padding, the output dimension becomes (24,24). Hence the size of the flattened layer changes as in Figure \ref{linearMorphism}. Due to this change, we must make the immediate following Fully Connected layer (FC layer) have a reduced number of neurons. 

In our framework, we use two Fully Connected Layers for classification. The first one, Variable FC Layer, is connected to the flattened NASGraph matrix and is morphed to reduce in size dynamically. The number of neurons in this Variable FC Layer is equal to the number of neurons in the Flattened Layer. We can do this because the image size of the CIFAR-10 dataset is (32,32). Thus, the maximum number of neurons in the Flatten layer can be $32*32 = 1024$. So now Variable FC also has $1024$ neurons. Therefore the size of the weight matrix becomes $1024*1024$ which is around 1 Million parameters, feasible to handle by today's computing standards. The next and the final layer has ten neurons corresponding to the ten classes in the CIFAR-10 dataset. Therefore, any changes in the NASGraph will be held until the Variable FC layer and we can have a constant size final Fully Connected layer for Classification.

\subsection{Process Flow}
\begin{figure}[t]
\centering
\includegraphics[width=0.3\textwidth,scale=0.6]{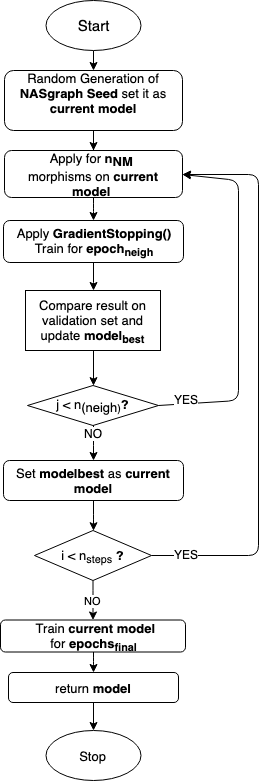}
\caption{Flowchart of our Iterative Hill Climbing technique for Neural Architecture Search} \label{hill_climbing}
\end{figure}

The flowchart in Fig. \ref{hill_climbing} shows the steps of our algorithm in a nutshell. The graphical architecture of the model is referred to as a NASGraph. First, a random architecture is generated to be the starting point of the state space search, called the NASGraph seed. The Hill climbing operation is carried for $n_{steps}$. In each step, $n_{neigh}$ neighbors are generated for the current best model. To create a neighbor, $n_{NM}$ morphism operations are applied as described in section \ref{Morphism}, \ref{Selection of Nodes For Morphism} and \ref{Random Selections}. The resultant neighbors imbibe the properties of the parent model along with having new attributes. These neighbors are then trained for $epoch_{neigh}$. Finally, the current best model is updated by comparing with the results of the neighbors on the validation dataset. After $n_{steps}$, the best model is trained for $epoch_{final}$ and the trained model is reported as the output of the algorithm.

\section{EXPERIMENTAL RESULTS}
\subsection{Dataset Overview}
For neural architecture search, we use CIFAR-10 dataset \cite{b29}. Convolution Neural Networks have shown to be very effective in working on the image problem domains. Thus, we showcase our results also on image datasets. To test how well does the architecture generated by our method on CIFAR dataset performs on images of a different domain, we use the handwritten digits MNIST dataset. Baseline results on Neural Architecture Search are usually on CIFAR / MNIST datasets. Therefore, a comparison of our method along with other methods is suitable for the same.

\subsection{Discussion on Results}

\begin{table}[h]
\caption{\label{tab:resultsTable}Results Comparison with different weight initialization techniques.
}
\centering
\begin{tabular}{|l|l|l|}
\hline
\multicolumn{1}{|c|}{\textbf{Learning Rate}}                        & \textbf{Xavier Initialization}               & \multicolumn{1}{c|}{\textbf{0/1 Initialization}} \\ \cline{2-3} 
                                                                    & \multicolumn{1}{c|}{Time (hrs) / Error (\%)} & \multicolumn{1}{c|}{Time (hrs) / Error (\%)}     \\ \hline
\begin{tabular}[c]{@{}l@{}}SGDR + \\ Gradient Stopping\end{tabular} & 19.4 / 4.96                                  & 19.6 / 5.2                                       \\ \hline
SGDR                                                                & 23.4 / 5.8                                   & 23.3 / 5.8                                       \\ \hline
\end{tabular}
\end{table}

For similarities in our methods, we choose the NASH \cite{b19} as our baseline. 

Parameters for NASH used to compare the results are as provided in \cite{b19}. The set values are given in Table \ref{tab:variableValues}. NASH requires us to feed a simple seed architecture which they mentioned as : Conv-Max pool-Conv-Max pool-Conv-FC-Softmax where Conv = Conv + BatchNorm + ReLU. This architecture has already been trained for 20 epochs. Using these parameters, the best possible error rate achieved by NASH is 5.2\% in one day of GPU training. 

Table \ref{tab:variableValues} also mentions the value of the parameters we have used, which is the same for all our architectures presented in this paper. We explore the same number of neighbors at any depth of the hill climbing as that of NASH. However, the exploration depth is higher in our case which in turn enables us to apply a higher number of morphisms on our network, and reach more depth, without requiring much time.  

We first explain the parameters used as follows.

\begin{enumerate}
    \item $n_{steps}$ is the depth of the hill climbing tree. 
    \item $n_{NM}$  is the number of times the morphism operation is applied on a hill climbing node.
    \item $n_{neigh}$ is the number of children generated in the hill climbing tree.
    \item $epoch_{neigh}$ is the number of epochs for which each child is trained for.
    \item $epoch_{final}$ is the number of epochs for which the final $model_{best}$ is trained.
    \item $\lambda_{start}$ and $\lambda_{end}$ are the parameters required for SGDR.
\end{enumerate}

\begin{table}[]
\caption{\label{tab:variableValues}Parameters for NASH and Our Method.}
\centering
\begin{tabular}{|l|l|l|}
\hline
\textbf{Variable}        & \textbf{NASH-2} & \textbf{Ours} \\ \hline
$n_{steps}$       & 8    & 10   \\
$n_{NM}$            & 5    & 5 \\
$n_{neigh}$       & 8    & 8    \\
$epoch_{neigh}$   & 17   & 16   \\
$epoch_{final}$   & 100  & 64   \\
$\lambda_{start}$ & 0.05 & 0.1  \\
$\lambda_{end}$   & 0.0  & 0.0  \\ \hline
\end{tabular}

\end{table}

\begin{table}[t]
\caption{\label{tab:otherMethods}Comparison of Neural Architecture Search on CIFAR-10 on different models.
}
\centering
\begin{tabular}{|l|l|l|}
\hline
\textbf{Model}                          & \textbf{Resource} & \textbf{Error(\%)} \\ \hline
Shake-Shake \cite{b11} & 2 days, 2 GPU     & 2.9                \\ \hline
RL-NAS \cite{b17} & ? days, 800 GPUs & 3.65 \\ \hline
NASH-1                                  & 0.5 days, 1 GPU   & 5.7                \\ \hline
NASH-2                                  & 1 day, 1 GPU      & 5.2                \\ \hline
Ours                                & 19.4 hours, 1 GPU & 4.96               \\ \hline
\end{tabular}

\end{table}

Since the objective of this paper is to propose a newer faster baseline, we must focus our scrutiny on three areas primarily, amount of resources used, the time required for the complete training process, and the final error percentage. In all the three fronts we either match the NASH or are better. Table \ref{tab:otherMethods} gives a comparison of our framework (with Linear Morphisms, Updated Operations \& Gradient Stopping with SGDR) with the two NASH models proposed in \cite{b19}, the Shake-Shake \cite{b11} model, and a reinforcement learning based Neural Architecture Search Scheme \cite{b17}. NASH-1 has $n_{steps} = 5$, while NASH-2 parameters are given in Table \ref{tab:variableValues}. Shake-Shake \cite{b11} is a manually crafted state-of-the-art model in CIFAR-10 classification which uses 2 GPUs and gives 2.9\% error in 2 days. However, on comparing with a Neural Architecture Search scheme like that of \cite{b17}, that although gives a low error of 3.65\%, but uses a mammoth 800 GPU system, is quite unrealistic for practical purposes. Our baseline, NASH scheme, uses 1 GPU and one day of training time to give a 5.2\% error, against our training time of 19.4 hours to produce better results at 4.96\% error. The reported error for our model is the average of three full runs on CIFAR-10 classification which yielded 5.1\%, 4.9\%, 4.9\% errors respectively averaging 19.4 hours of training time. We also note that our method searches for comparatively more of the neural architecture space in less time and produces better accuracy.

\subsection{On a Different Dataset}

We also performed classification using the architectures produced with CIFAR-10, on the MNIST dataset. It was done to gain insights regarding how the architecture performs on other image datasets. Low error percentages by architectures generated from one type of dataset, upon another dataset, would mean either similarity in the dataset, or the robustness of the architecture itself. Since MNIST and CIFAR-10 are known to be very different datasets, where MNIST consists of handwritten digits, and CIFAR-10 consists of images of various real-life objects, we chose MNIST for this purpose. Upon using the same architecture, for which the results on CIFAR-10 are mentioned in Tables \ref{tab:resultsTable} and \ref{tab:otherMethods}, we obtain 0.28\% error on MNIST dataset. The current state-of-the-art on MNIST is Rmdl: Random multimodel deep learning for classification \cite{b22} which boasts an error percentage of 0.18. However, our architecture would still stand in the top 10\footnote{https://paperswithcode.com/sota/image-classification-mnist-handwritten-d} error percentages reported until June 2018.


\section{Conclusion \& Future Work}
We introduced a Neural Architecture Search in Hill Climbing domain using morphism operations that use our novel Gradient Update scheme. Our method is not just fast and uncomplicated, but it also outperforms many existing automated architecture search methods with lesser GPU days and resources. Our experiments performed on the CIFAR-10 and MNIST datasets and yielded satisfactory results.

Our approach can quite comfortably deliver as a basis for the development of more sophisticated techniques that yield further improvements in the existing performance. Certain operations are being executed randomly in our method, such as the selection of a morphism operation or selection of a node for performing the morphism steps. One can make these selections intelligently by using reinforcement learning or genetic algorithms. Essential criteria for selection of nodes would be to pick convolution blocks from those regions of the network which have been having significant changes in their weights in the recent iterations. Again, there are different ways to initialize the weights in the morphism operations. We experimented with two of many possibilities, and expect further research in weight initialization would yield better results.

Apart from improvements in the existing work, we find research in Learning Curve Extrapolation for dynamic neural networks in its initial stages, but would immensely speed up the overall hill climbing iterative algorithm. However, it is currently there for static \cite{b26, b27} neural network graphs.


\appendix[Sample Output of NASGraph]
\label{appendix}

\begin{figure*}[htbp]
\centering
\includegraphics[width=0.35\textwidth]{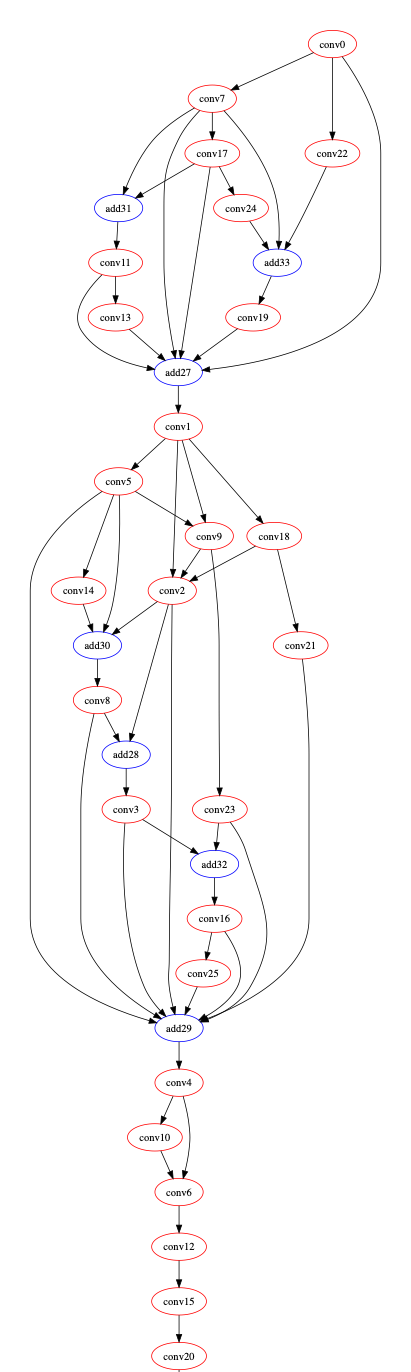}
\caption{Sample output of NASGraph on CIFAR-10 dataset} \label{samplegraph}
\end{figure*}

Fig. \ref{samplegraph} shows a sample graph produced by the NASGraph using the parameters mentioned in Table II. The generated network can be seen to have convolution blocks as represented by red bubbles, and Merge layers (which can be either the Merge or the Add function) represented by the blue ’add’ bubbles. The graph clearly holds the Topological Ordering condition as mentioned a must for the NASGraph. In addition, it is seen to have Skip connections for example node conv0 is connected to add27 providing input to conv1, along with conv13, conv19, conv7, conv17, conv11. The NASGraph can generate quite complicated neural networks and provide good accuracy rates on CIFAR-10.

\begin{thebibliography}{00}
\bibitem{b1} He, K., Zhang, X., Ren, S., \& Sun, J. (2016). Deep residual learning for image recognition. In Proceedings of the IEEE conference on computer vision and pattern recognition (pp. 770-778).
\bibitem{b2} Karpathy, A., \& Fei-Fei, L. (2015). Deep visual-semantic alignments for generating image descriptions. In Proceedings of the IEEE conference on computer vision and pattern recognition (pp. 3128-3137).
\bibitem{b3} Silver, D., Huang, A., Maddison, C. J., Guez, A., Sifre, L., Van Den Driessche, G., ... \& Dieleman, S. (2016). Mastering the game of Go with deep neural networks and tree search. nature, 529(7587), 484.
\bibitem{b4} Sutskever, I., Vinyals, O., \& Le, Q. V. (2014). Sequence to sequence learning with neural networks. In Advances in neural information processing systems (pp. 3104-3112).4
\bibitem{b5} Schaffer, J. D., Whitley, D., \& Eshelman, L. J. (1992, June). Combinations of genetic algorithms and neural networks: A survey of the state of the art. In [Proceedings] COGANN-92: International Workshop on Combinations of Genetic Algorithms and Neural Networks (pp. 1-37). IEEE.
\bibitem{b6} Liu, H., Simonyan, K., Vinyals, O., Fernando, C., \& Kavukcuoglu, K. (2017). Hierarchical representations for efficient architecture search. arXiv preprint arXiv:1711.00436.
\bibitem{b7} Miikkulainen, R., Liang, J., Meyerson, E., Rawal, A., Fink, D., Francon, O., ... \& Hodjat, B. (2019). Evolving deep neural networks. In Artificial Intelligence in the Age of Neural Networks and Brain Computing (pp. 293-312). Academic Press.
\bibitem{b8} Real, E., Aggarwal, A., Huang, Y., \& Le, Q. V. (2018). Regularized evolution for image classifier architecture search. arXiv preprint arXiv:1802.01548.
\bibitem{b9} Real, E., Moore, S., Selle, A., Saxena, S., Suematsu, Y. L., Tan, J., ... \& Kurakin, A. (2017, August). Large-scale evolution of image classifiers. In Proceedings of the 34th International Conference on Machine Learning-Volume 70 (pp. 2902-2911). JMLR. org.
\bibitem{b10} Suganuma, M., Shirakawa, S., \& Nagao, T. (2017, July). A genetic programming approach to designing convolutional neural network architectures. In Proceedings of the Genetic and Evolutionary Computation Conference (pp. 497-504). ACM.
\bibitem{b11} Gastaldi, X. (2017). Shake-shake regularization. arXiv preprint arXiv:1705.07485.
\bibitem{b12} Zoph, B., \& Le, Q. V. (2016). Neural architecture search with reinforcement learning. arXiv preprint arXiv:1611.01578.
\bibitem{b13} Williams, R.J.: Simple statistical gradient-following algorithms for connectionist
reinforcement learning. Mach. Learn. 8(3-4) (May 1992) 229–256
\bibitem{b14} Zoph, B., Vasudevan, V., Shlens, J., \& Le, Q. V. (2018). Learning transferable architectures for scalable image recognition. In Proceedings of the IEEE conference on computer vision and pattern recognition (pp. 8697-8710).
\bibitem{b15} Cai, H., Chen, T., Zhang, W., Yu, Y., \& Wang, J. (2018, April). Efficient architecture search by network transformation. In Thirty-Second AAAI Conference on Artificial Intelligence.
\bibitem{b16} Baker, B., Gupta, O., Naik, N., \& Raskar, R. (2016). Designing neural network architectures using reinforcement learning. arXiv preprint arXiv:1611.02167.

\bibitem{b17} Bello, I., Zoph, B., Vasudevan, V., \& Le, Q. V. (2017, August). Neural optimizer search with reinforcement learning. In Proceedings of the 34th International Conference on Machine Learning-Volume 70 (pp. 459-468). JMLR. org.
\bibitem{b18} Wei, T., Wang, C., Rui, Y., \& Chen, C. W. (2016, June). Network morphism. In International Conference on Machine Learning (pp. 564-572).
\bibitem{b19} Elsken, T., Metzen, J. H., \& Hutter, F. (2017). Simple and efficient architecture search for convolutional neural networks. arXiv preprint arXiv:1711.04528.
\bibitem{b20} Ruder, S. (2016). An overview of gradient descent optimization algorithms. arXiv preprint arXiv:1609.04747.
\bibitem{b21} Miller, G. F., Todd, P. M., \& Hegde, S. U. (1989, June). Designing Neural Networks using Genetic Algorithms. In ICGA (Vol. 89, pp. 379-384).

\bibitem{b22} Kowsari, K., Heidarysafa, M., Brown, D. E., Meimandi, K. J., \& Barnes, L. E. (2018, April). Rmdl: Random multimodel deep learning for classification. In Proceedings of the 2nd International Conference on Information System and Data Mining (pp. 19-28). ACM.
\bibitem{b23} Loshchilov, I., \& Hutter, F. (2016). Sgdr: Stochastic gradient descent with warm restarts. arXiv preprint arXiv:1608.03983.
\bibitem{b24} Ioffe, S., \& Szegedy, C. (2015). Batch normalization: Accelerating deep network training by reducing internal covariate shift. arXiv preprint arXiv:1502.03167.

\bibitem{b25} Simonyan, K., \& Zisserman, A. (2014). Very deep convolutional networks for large-scale image recognition. arXiv preprint arXiv:1409.1556.

\bibitem{b26}Klein, A., Falkner, S., Springenberg, J. T., \& Hutter, F. (2016). Learning curve prediction with Bayesian neural networks.

\bibitem{b27}Domhan, T., Springenberg, J. T., \& Hutter, F. (2015, June). Speeding up automatic hyperparameter optimization of deep neural networks by extrapolation of learning curves. In Twenty-Fourth International Joint Conference on Artificial Intelligence.

\bibitem{b28} Paszke, A., Gross, S., Chintala, S., Chanan, G., Yang, E., DeVito, Z., ... \& Lerer, A. (2017). Automatic differentiation in pytorch.

\bibitem{b29}
Krizhevsky, A., \& Hinton, G. (2009). Learning multiple layers of features from tiny images (Vol. 1, No. 4, p. 7). Technical report, University of Toronto.

\end{thebibliography}
\end{document}